\documentclass{article}

\PassOptionsToPackage{numbers}{natbib}

\usepackage[preprint]{neurips_2020}

     \usepackage{neurips_2020}

\usepackage[utf8]{inputenc} 
\usepackage[T1]{fontenc}    
\usepackage{hyperref}       
\usepackage{url}            
\usepackage{booktabs}       
\usepackage{amsfonts}       
\usepackage{nicefrac}       
\usepackage{microtype}      
\usepackage[pdftex]{graphicx}
\DeclareUnicodeCharacter{FB01}{fi}

\title{Channel Distillation: Channel-Wise Attention for Knowledge Distillation}

\author{
	Zaida Zhou\textsuperscript{1}\quad\quad Chaoran Zhuge\textsuperscript{1}\quad\quad   Xinwei Guan\textsuperscript{1}\quad\quad   Wen Liu\textsuperscript{2} \\
	\textsuperscript{1}University of Science and Technology of China\quad\quad   \textsuperscript{2}Sun Yat-sen University \\
	\texttt{\{zdzhou01, zgcr, xinaruto\}@mail.ustc.edu.cn}\quad\quad   \texttt{liuw256@mail2.sysu.edu.cn} \\
}

\begin{document}

\maketitle

\begin{abstract}
  Knowledge distillation is to transfer the knowledge from the data learned by the teacher network to the student network, so that the student has the advantage of less parameters and less calculations, and the accuracy is close to the teacher. In this paper, we propose a new distillation method, which contains two transfer distillation strategies and a loss decay strategy. The first transfer strategy is based on channel-wise attention, called Channel Distillation (CD). CD transfers the channel information from the teacher to the student. The second is Guided Knowledge Distillation (GKD). Unlike Knowledge Distillation (KD), which allows the student to mimic each sample's prediction distribution of the teacher, GKD only enables the student to mimic the correct output of the teacher. The last part is Early Decay Teacher (EDT). During the training process, we gradually decay the weight of the distillation loss. The purpose is to enable the student to gradually control the optimization rather than the teacher. Our proposed method is evaluated on ImageNet and CIFAR100. On ImageNet, we achieve 27.68\% of top-1 error with ResNet18, which outperforms state-of-the-art methods. On CIFAR100, we achieve surprising result that the student outperforms the teacher. Code is available at \url {https://github.com/zhouzaida/channel-distillation}.

\end{abstract}

\section{Introduction}

In recent years, deep learning has made great progress in computer vision and natural language processing. The emergence of the convolutional neural network AlexNet \cite{ref1} has caught the attention of many scholars, and it far exceeds traditional algorithms in image classification. Researches on convolutional neural networks have found that a deeper network can extract more abstract semantic information, so the neural network becomes deeper and deeper, and the network's representation ability is getting better and better. However, the deeper network will be difficult to converge, and the gradient disappears when it performs back-propagation \cite{ref22, ref23}. The proposal of the ResNet \cite{ref2} and the BN \cite{ref3} have solved this problem to a certain extent, but a large number of network parameters and calculations are unacceptable on the mobile terminal for real-time inference. Knowledge distillation is a universal solution for model compression. Generally, we first train a large teacher network, and then use the teacher network to supervise the training of small student network, thereby the performance of the student network is improved and the student network can be served at a lower cost.

\begin{figure}
  \centering
  \includegraphics[width=3.5in]{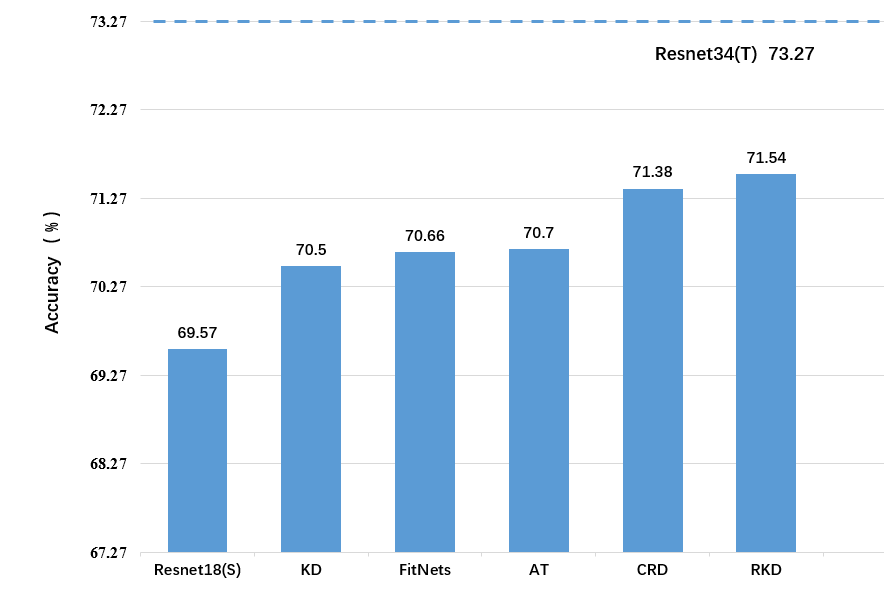}
  \caption{The different distillation methods with Resnet34 as the teacher and Resnet18 as the student. Obviously, the accuracy of the teacher is 73.27\%, which is about 2\% higher than the current distillation method.}
  \label{baseline}
\end{figure}

In order to achieve a better knowledge distillation, many methods have been proposed in \cite{ref4, ref5, ref6, ref7, ref8, ref14, ref20}. Geoffrey Hinton et al. \cite{ref13} used Knowledge Distillation (KD) loss to minimize the difference between the teacher's output distribution and the student's output distribution. Adriana Romero et al. \cite{ref14} used L2 loss to make the feature map of the student simulate the feature map of the teacher. Byeongho Heo et al. \cite{ref7} utilized the decision boundary of the teacher to guide the student. However, these methods still cannot transfer the knowledge effectively. Although the student obtains the knowledge transferred by the teacher, there is still a certain gap in performance between them, as shown in Figure~\ref{baseline}. There are three main reasons for this. One is that the knowledge transferred from the teacher is not good enough, and the student cannot accurately learn the essential information from the teacher. Second is that the prediction of the teacher is not completely correct, during training, if the student makes a decision with reference to the decisive result of the teacher, the poor output of the teacher will have a bad influence on the student instead. Additionally, there is a margin between the teacher and the student since they have different structure, which will make the student unable to find its own optimization space if we always let the teacher supervise it.

To solve the three problems mentioned above, inspired by \cite{ref10, ref11, ref12, ref18}, we propose a novel method of knowledge distillation. We transfer the knowledge to the student by the method of Channel-Wise Distillation (CD), which is a special attention we will explain in detail in Section~\ref{cd}, so that the student can extract feature more effectively. At the same time, to avoid the negative impact of the teacher on the student, we propose Guided Knowledge Distillation (GKD) based on KD, only using the correct output of the teacher as knowledge to guide the student, that is, only when the teacher predicts correctly, the student simulate the output of the teacher. In addition, we also propose Early Decay Teacher (EDT) to reduce the proportion of distillation loss during the training, making sure the student is able to find its own optimization space.

To summarize, the contributions of our work can be listed as follows:

\begin{itemize}
\item Different to other knowledge distillations, we let the student learn the teacher's ability to
recognize channel representation
\item We only calculate the loss of samples classified correctly by the teacher to guide the student
\item The supervision on student will be gradually decreased in late period of the training process
\item We proved our method has reached state-of-the-art performance, which is illustrated in Table~\ref{sota-table}
\end{itemize}

\section{Related Work}

{\bf Knowledge distillation methods.} Caruana et al. \cite{ref27} first proposed a method to compress the functions learned by the large model into smaller and faster model which can match the results of the larger model. Later, Hinton et al. \cite{ref13} applied this technology to help the teacher transfer the output distribution as a form of knowledge to the student, so that the output distribution of the student was close to the teacher. Compared with directly use of the one-hot hard label, imitating this soft distribution can also enable the student to learn more inter-class information. Adriana Romero et al. \cite{ref14} found that such kind of constraint is not sufficient at the label level. The deeper the network, the harder it is to transmit the final layer of supervisory signals back to the front. His FitNets \cite{ref14} added some supervisory signals in the middle of the network, which utilized hint training to constrain that the output of the middle layer of two models should be as close as possible. \cite{ref8, ref21} focus on batch dimensions and reduce spatial information, using a method to pass the correlation between samples, similar to gram matrix (Gatys, Ecker, and Bethge 2016) \cite{ref19}. Sergey Zagoruyko and Nikos Komodaki proposed Attention Transfer \cite{ref9}, which uses spatial-attention to transfer the attention information of the teacher model to the student model in the intermediate layer. Yim et al. \cite{ref5} defined a Flow of Solution Procedure (FSP) matrix of the feature map between the middle layer, which represents the way that the teacher processes the information between the two layers, and this processing way is passed to the student by distillation.

{\bf Attention.} The attention mechanism stems from the study of human vision. Owing to the bottleneck of information processing, humans will selectively pay attention to part of the information while ignoring others. In the field of computer vision, attention mechanisms are put in place to perform visual information processing \cite{ref28, ref29, ref30, ref31, ref32, ref33}, such as local image feature extraction, saliency detection, sliding window method, etc. All of them can be considered as attention mechanisms. Attention mechanism is typically used in the processing of feature maps. CBAM \cite{ref15}, proposed by Sanghyun Woo et al., applies attention on both channel and spatial position in the feature map which considers global average pooling (GAP) and global max pooling (GMP) information. Jie Hu et al. proposed SE-NET \cite{ref10}, which utilizes GAP and fully connected (FC) to obtain the weight on the channel. SKNET \cite{ref16} convolves the feature map with $3 \times 3$ and $5 \times 5$ kernels respectively, and then obtains the weight of the two convolution results through GAP to do attention. SGE-NET \cite{ref17} groups the channels of the  feature map and uses GAP in each group to get a global vector, $ g $. Next, $ g $ is utilized to multiply the original feature map by position-wise dot, and then the normalization operation is performed to get mask $ a $. Finally, $sigmoid(a)$ is used to do attention.

{\bf Early stop.} Most of the research on early stop is limited to a single model. The researchers observed that when the model is over-trained, poor generalization will occur. This phenomenon is called network overfitting, therefore stopping training ahead of time is a general technique for solving the poor generalization of the model \cite{ref22}. Maren Mahsereci et al. \cite{ref23} proposed a novel early stopping criterion based on fast-to-compute local statistics of the computed gradients. Rich Caruana \cite{ref26} also verified that early stopping combined with back-propagation is so effective that large networks can be trained without significant overfitting. Although early stopping technology is ued infrequently in the field of distillation, Jang Hyun Cho and Bharath Hariharan \cite{ref24} had observed a phenomenon in their experiments that not always teachers with better performance can teach students better. When the gap (network structure, network capacity) between the teacher and the student is huge, the student isn't able to mimic well, so they apply the early stop method to eliminate the effect of distillation loss on the total loss.

\section{Method}
\label{method}

In this section, we first introduce the insight behind the Channel Distillation, then we explain what kind of intuition we are based on to improve Knowledge Distillation and design Guided Knowledge Distillation, and finally describe our loss decay strategy, Early Decay Teacher.

\subsection{Channel Distillation}
\label{cd}

Our method derives from the idea of SENet \cite{ref10}. In SENet, Channel-Wise Attention enables the model to learn the weight of each channel, and then multiplies the weight back to the original channel. The features of those important channels are enhanced, and the insignificant are weakened, so that the extracted features are more directional and the network predicts better. The weight of each channel is given by:

\begin{equation}
w_c = \frac{1}{H \times W}\sum_{i=1}^{H}\sum_{h=1}^{W}u_c(i,j)
\label{eq:wc}
\end{equation}

where $ w_c $ is the weight of $ c^{th} $ channel. $ H, W $ are the spatial dimensions of the feature map and $ u_c(i,j) $ is the activation.

Each channel of the feature map corresponds to a visual pattern \cite{ref38}, but the importance of each channel's visual pattern is different. Since teacher's performance is better than student, we think that the visual pattern learned by teacher is more accurate, and we want student to learn the teacher's visual pattern. We use GAP to calculate the importance of each channel's feature map, which represnets attention information of each channel. Then we consider the attention information of each channel's feature map as knowledge. The student and the teacher will calculate the  attention information of each channel from their feature maps respectively, and then the teacher will supervise the student to learn the  attention information of each channel, and transfer the attention information to the student. In this way, the student can learn the teacher's  attention information of each channel, thereby improving its performance. Normally, the number of layers of the teacher and the student is inconsistent. For simple processing, we only perform Channel Distillation between the networks where the spatial resolution decreases. In addition, if the number of channels is mismatch, we follow the method of FitNets \cite{ref14} which uses $ 1\times1 $ convolution to upgrade the dimension. The student's feature map is first upgraded to the same number of channels as the teacher, and then Channel Distillation is performed. The formulation of CD loss is deﬁned as:

\begin{equation}
CD(s,t) = \frac{\sum_{i=1}^{n}\sum_{j=1}^{c}{(w_s^{ij}-w_t^{ij})}^2}{n \times c}
\label{eq:cd}
\end{equation}

where $ CD(s,t) $ means the CD loss between the student and the teacher. $ w_{ij} $ is the weight of $ j^{th} $ channel of $ i^{th} $ sample. $ c $ 
represents the number of channels.

\begin{figure}
	\centering
	\includegraphics[width=4in]{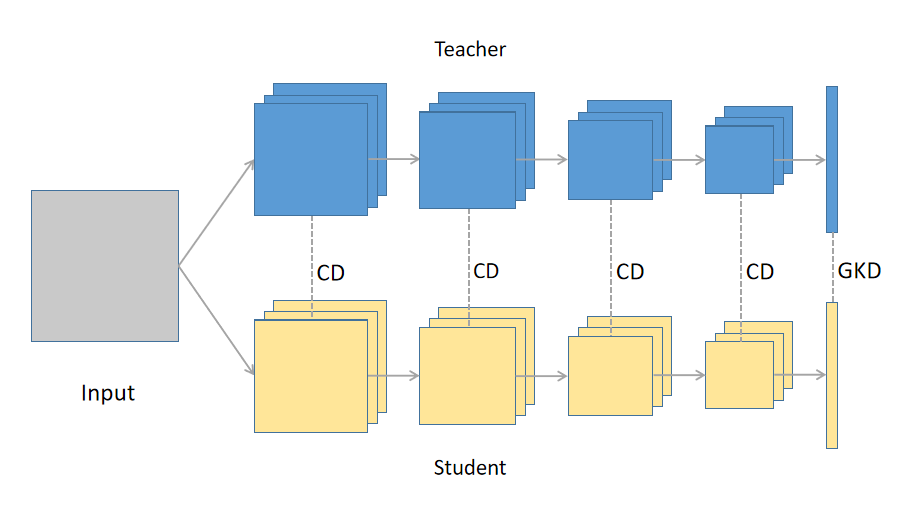}
	\caption{The architecture of our distillation network, where CD means Channel Distillation and GKD means Guided Knowledge Distillation.}
	\label{arch}
\end{figure}

Figure~\ref{arch} is the structure of our distillation network. The upper blue part represents the teacher network and the lower red part represents
 the student network. Among them, the four squares in the front represent the feature map, the vertical bar in the back represents the predition
 after softmax, CD represents the channel distillation above, and GKD represents the Guided Knowledge Distillation. We will introduce GKD in the following section.

\subsection{Guided Knowledge Distillation}
\label{gkd}

Our Guided Knowledge Distillation (GKD) is devised on the basis of KD \cite{ref13}. The basic idea of KD is to calculate the prediction distribution between the teacher and the student. By gradually minimizing the divergence between them, the output distribution of the student is similar to the teacher's. The formula of KD is given by:
 
\begin{equation}
p=softmax(\frac{a}{T})
 \label{eq:p}
\end{equation}
\begin{equation}
KD(s,t)=\frac{\sum_{i=1}^{n}KL(p_s^i,p_t^i)}{n}
\label{eq:kd}
\end{equation}

where $ p $ is probability distribution computed by logit, $ a $. And $ T $ is a temperature. Utilizing a higher value for $ T $ produces softer probability distribution over classes. $ n $ is the batch size. In addition, the $ KD(s,t) $ is the mean of Kullback-Leibler divergence (KL) between $ p_s^i $ and $ p_t^i $.

Although the teacher network is more accurate than the student, the teacher still has some prediction errors in a batch, so there will be a serious
 problem. When the teacher predicts incorrectly, the knowledge will also be transferred to the student, which will deteriorate the performance of the student. 
Therefore, we improved on the basis of KD to get GKD. Our intention is that the teacher only transfer the positive predition distribution to the student and directly 
ignore the negative. More specifically, our approach is to only backward the KD loss of those samples correctly predicted by the teacher network and ignore incorrectly. 
As shown in Figure~\ref{arch}. The formula of GKD is defined as follows:

\begin{equation}
GKD(s,t)=\frac{\sum_{i=1}^{n}I(p_t^i,y_i)KL(p_s^i,p_t^i)}{\sum_{i=1}^{n}I(p_t^i,y_i)}
\label{eq:gkd}
\end{equation}

where $ I $ is an indicator function and $ I(p_t^i,y_i) $ is $1$ when the output of the teacher equals true label, or else $ I(p_t^i,y_i) $ is $0$. For example, supposing a batch size has $ n $ samples and the teacher correctly predicts $ n_1 $ samples, we only calculate the GKD loss of the $ n_1 $ samples.

\subsection{Early Decay teacher}
\label{edt}

\cite{ref24} proposed that the impact of distillation is not always positive. In the early stage of network
 training, KD helps student train, but it will inhibit student learning better in the later, so at the appropriate time stopping the teacher's 
supervision helps the student train. The experimental results show that at a certain epoch, the loss of cross-entropy will reversely rise. At this 
point, it will be better to stop the teacher's supervision, but the practice of directly stopping the distillation is hard. Therefore, we propose a 
relatively soft approach. Our distillation loss weight decreases as the learning rate decreases. The formula is defined as follows:

\begin{equation}
EDT(\alpha)=\alpha \times \lambda^{n_e/n}
\label{eq:edt}
\end{equation}

where $ \alpha $ is initial weight of distllation loss, and $ \lambda $ is a constant coefficient. $ n_e $ denotes $ n^{th} $ epoch in the entire training process, and $ n $  is an empirical value representing the number of epochs we reduce our weight of loss.

We only decrease the weight of CD loss. For the GKD loss, we don't decrease its weight throughout the training process, because the GKD loss only teaches the correct knowledge from the teacher to the student.

In the end, the student network is then trained by optimizing the following loss function:

\begin{equation}
Loss(s,t)=EDT(\alpha)CD(s,t)+GKD(s,t)+CE(s,y)
\label{eq:loss}
\end{equation}

where $ CE $ is the cross entropy loss function commonly used in classification and $ y $ denotes the true label.

\section{Experiments}
\label{experiments}

We evaluated the efficiency of our distillation method on several datasets and networks. Since most of other distillation methods reported the performance in this domain, we also compared our results with other distillation methods. All experiments were conducted using PyTorch \cite{ref37} framework.

\subsection{Datasets}

We validated our distillation method on CIFAR100 \cite{ref34} and ImageNet \cite{ref35} datasets. CIFAR100 contains 50,000 training images and 10,000 test images 
with 100 categories. ImageNet provides 1.2 million images from 1,000 classes for training and 50,000 images for validation.

\subsection{Implement Details}
\label{implement-detail}

First, we reproduced the results of distillation using cross-entropy loss and KD loss. Subsequently, we conducted an ablation study in Section~\ref{ablation-study}
to validate the effectiveness of the CD loss, GKD loss, and EDT methods proposed in Sections~\ref{cd}, \ref{gkd}, and \ref{edt}, respectively.
Our teacher network and student network use ResNet family network. During CIFAR100 and ImageNet training, the teacher network is initialized with 
the pre-training weights on the training set.

For all experiments on ImageNet, we used stochastic gradient descent (SGD) optimizer with momentum 0.9 and weight decay 1e-4 for training. For the
 teacher network, we froze the parameters weight of all teacher layers. The input size is set to $ 224\times224 $. The initial learning rate is 0.1, divided
 by 10 at 30, 60 and 90 epochs. We set batch size to 256 and train for 100 epochs. We used a standard data augmentation scheme with crop and flip, and
 normalize the input images using the channel means and standard deviations.

When training on CIFAR100, the values of the training parameters are different. Since the image size of the CIFAR100 dataset is $ 32\times32 $, in order
 to make the ResNet network learn better features on low resolution images, we modified the first convolution layer parameters of ResNet from 
kernel\_size = 7, stride = 2 and padding = 3 to kernel\_size = 3, stride = 1 and padding = 1. Then we removed the max pooling layer after the
 first convolution layer. SGD optimizer are still used, but the weight decay is set to 5e-4 and batch size is set to 128. The initial learning
 rate is still 0.1, divided by 5 at 60, 120 and 160 epochs. The entire model is trained with 200 epochs. For CIFAR100 dataset, we used padding=4, 
flip and crop to augment input data, and normalized the input images with the channel means and standard deviations.

\subsection{Ablation Study}
\label{ablation-study}

\begin{table}
	\caption{Top error on ImageNet validation set (\%). All experiments follow the training settings in Section~\ref{implement-detail}.}
	\label{imagenet-table}
	\centering
	\begin{tabular}{lllll}
		\toprule
		Method   & Model & Top-1 error(\%) & Top-5 error(\%) \\
		\midrule
		teacher & ResNet34 & 26.73 & 8.74    \\
		student & ResNet18 & 30.43 & 10.76     \\
		KD & ResNet34-ResNet18 & 29.50 & 9.52 \\
		{\bf CD(our)} & ResNet34-ResNet18 & 28.53 & 9.56 \\
		{\bf CD+GKD(our)} & ResNet34-ResNet18  & 28.26 & 9.41 \\
		{\bf CD+GKD+EDT(our)} & ResNet34-ResNet18 & 27.61 & 9.2 \\
		\bottomrule
	\end{tabular}
\end{table}

In this part, we conducted three ablation studies to validate the effectiveness of the CD loss, GKD loss, and EDT methods proposed in Sections~\ref{cd}, \ref{gkd}, and \ref{edt}, respectively. On ImageNet dataset, we mainly use ResNet34 as the teacher network and ResNet18 as the student network. In Table~\ref{imagenet-table}, we report our experiment results on ImageNet.

Obviously, when only using CD loss distillation, the performance of the student network ResNet18 has exceeded the performance of the student 
network ResNet18 using KD loss distillation. This shows that our proposed CD loss make the student network learn more knowledge 
from the teacher network. Next, we add GKD loss on the basis of CD loss, speed up the student's learning process of positive knowledge, and get better performance. We continue to add the EDT strategy and gradually reduce the supervision of CD loss in the later stages of training so that student learns  a better local optimum, which makes student even better.

\begin{table}
	\caption{Top error on CIFAR100 validation set (\%). All experiments follow the training settings in Section~\ref{implement-detail}.}
	\label{cifar-table}
	\centering
	\begin{tabular}{llll}
		\toprule
		Method   & Model  & Top-1 error(\%) & Top-5 error(\%) \\
		\midrule
		teacher & ResNet152 & 19.09 &  4.45    \\
		student & ResNet50 & 22.02 & 5.74     \\
		KD & ResNet152-ResNet50 & 20.36 & 4.94 \\
		{\bf CD(our)} & ResNet152-ResNet50 & 20.08 & 4.78 \\
		{\bf CD+GKD(our)} & ResNet152-ResNet50 & 19.49 & 4.85 \\
		{\bf CD+GKD+EDT(our)} & ResNet152-ResNet50 & 18.63 & 4.29 \\
		\bottomrule
	\end{tabular}
\end{table} 

On CIFAR100 dataset, we mainly use ResNet152 as the teacher network and ResNet50 as the student network. Due to our improvement of ResNet network structure, 
our ResNet152 and ResNet50 baseline performance is much higher than the results reported in other papers. On this basis, we use ResNet152 teacher to distill
 ResNet50 student. In Table~\ref{cifar-table}, we report our experiment results on CIFAR100. On CIFAR100, our proposed method makes the performance of ResNet50 student exceed the ResNet152 teacher.

It is easy to find that on a small-capacity dataset such as CIFAR100, our method is 1.73\% higher than the KD loss. For the CIFAR100 dataset, ResNet152 has a large network capacity, so that ResNet152 network contains many useless parameters. In order to avoid teaching invalid knowledge to the student, our distillation method reduces the supervision of ResNet152 teacher in the later stages of training, and uses GKD losses only to teach correct knowledge to the student, which makes our ResNet50 student perform better than ResNet152 teacher.

\subsection{Performance comparison of different knowledge distillation methods}

In this part, we compared our method with other knowledge distillation methods (including KD, FitNets, AT, RKD, CRD) on the ImageNet dataset. 
In Table~\ref{sota-table}, we report the performance of different knowledge distillation methods on ImageNet.

\begin{table}
	\caption{Top-1 and Top-5 error rates (\%) of student network ResNet18 on ImageNet validation set. We compare our method with KD \cite{ref13} (Hinton et al., 2015), FitNets \cite{ref14} (Adriana Romero et al., 2015), AT \cite{ref9} (Sergey Zagoruyko et al., 2016), RKD \cite{ref6} (Mengya Gao et al., 2020), and CRD \cite{ref36} (Yonglong Tian et al., 2020).}
	\label{sota-table}
	\centering
	\begin{tabular}{lllll}
		\toprule
		Method   & Model  & GFLOPS(G)     & Top-1 error(\%) & Top-5 error(\%) \\
		\midrule
		teacher & ResNet34 & 3.672  & 26.73 & 8.74    \\
		student & ResNet18     & 1.820 & 30.43 & 10.76     \\
		KD & ResNet34-ResNet18 & 1.820 & 29.50 & 9.52 \\
		FitNets & ResNet34-ResNet18 & 1.820 & 29.34 & 10.77 \\
		AT & ResNet34-ResNet18 & 1.820 & 29.30 & 10.00 \\
		RKD & ResNet34-ResNet18 & 1.820 & 28.46 & 9.74 \\
		{\bf CD+GKD+EDT(our)} & ResNet34-ResNet18 & 1.820 & 27.61 & 9.2 \\
		\bottomrule
	\end{tabular}
\end{table}

It can be found from Table~\ref{sota-table} that using KD loss to distill on ImageNet can't achieve an excellent result, and only improve the accuracy 
of the student network by 0.93\% (top-1) and 1.24\% (top-5). The performance of  FitNets and AT is slightly better than KD loss, but the 
accuracy of top-1 is only improved by less than 0.2\% (top-1) than KD loss, and the accuracy of top-5 is even lower than KD loss. RKD 
improves the accuracy of student networks by 1.97\% (top-1) and 1.02\% (top-5), which is much better than KD, FitNets, AT and CRD methods. 
Our method is improved by 0.78\% (top-1) and 0.35\% (top-5) than RKD, which shows that our method make the student network learn knowledge 
from the teacher network more effectively.

\section{Conclusion}

We proposed Channel-Wise Attention for Knowledge Distillation: a novel form of knowledge distillation that aims to transfer the channel attention information from the teacher to the student, not just mimicking the teacher's representation in label. Our experiments demonstrate the effectiveness of our proposed method compared with other knowledge distillation methods. We have also shown that our CD+GKD+EDT method achieves the state-of-the-art on ImageNet and suprising result on CIFAR100. The method we proposed can be used for model compression, which can significantly reduce the model calculation cost.

\end{document}